\documentclass[letterpaper]{article} 
\usepackage{aaai2026}
\usepackage{times}  
\usepackage{helvet}  
\usepackage{amsmath}
\usepackage{amsfonts}
\usepackage{booktabs}
\usepackage{courier}  
\usepackage[hyphens]{url}  
\usepackage{graphicx} 
\def\UrlFont{\rm}  
\usepackage{natbib}  
\usepackage{caption} 
\usepackage{algorithm}
\usepackage{algorithmic}
\usepackage[most]{tcolorbox}
\usepackage{tabularx}
\usepackage{subcaption}
\usepackage{atbegshi}
\usepackage{tikz}

\usepackage{newfloat}
\usepackage{listings}
\DeclareCaptionStyle{ruled}{labelfont=normalfont,labelsep=colon,strut=off} 
\floatstyle{ruled}
\newfloat{listing}{tb}{lst}{}
\floatname{listing}{Listing}
\definecolor{rubinered}{HTML}{CE0058}

\definecolor{green}{rgb}{0.0, 0.65, 0.31}

\definecolor{bleudefrance}{rgb}{0.19, 0.55, 0.91}

\definecolor{ao(english)}{rgb}{0.0, 0.5, 0.0}

\definecolor{violet}{HTML}{6a51a3}

\definecolor{blue}{rgb}{0.0, 0.0, 1.0}

\title{AgentSense: Virtual Sensor Data Generation Using LLM Agents in Simulated Home Environments}

\author {
    Zikang Leng\textsuperscript{\rm 1}\equalcontrib,
    Megha Thukral\textsuperscript{\rm 1}\equalcontrib,
    Yaqi Liu\textsuperscript{\rm 1}\equalcontrib,
    Hrudhai Rajasekhar\textsuperscript{\rm 1},
    Shruthi K. Hiremath\textsuperscript{\rm 1},
    Jiaman He\textsuperscript{\rm 2},
    Thomas Plötz\textsuperscript{\rm 1}
}
\affiliations {
    \textsuperscript{\rm 1}School of Interactive Computing, Georgia Institute of Technology, Atlanta, USA\\
    \textsuperscript{\rm 2}RMIT University, Melbourne, Australia\\
    \{zleng7, mthukral3, yliu3387, hrajasekhar3, shiremath9, thomas.ploetz\}@gatech.edu, jiaman.he@student.rmit.edu.au
}

\newcommand{\equalcontrib}{*Equal contribution}

\begin{document}
\maketitle

\begin{abstract}
    \noindent
A major challenge in developing robust and generalizable Human Activity Recognition (HAR) systems for smart homes is the lack of large and diverse labeled datasets. Variations in home layouts, sensor configurations, and individual behaviors further exacerbate this issue. To address this, we leverage the idea of embodied AI agents—virtual agents that perceive and act within simulated environments guided by internal world models. We introduce AgentSense, a virtual data generation pipeline in which agents live out daily routines in simulated smart homes, with behavior guided by Large Language Models (LLMs). The LLM generates diverse synthetic personas and realistic routines grounded in the environment, which are then decomposed into fine-grained actions. These actions are executed in an extended version of the VirtualHome simulator, which we augment with virtual ambient sensors that record the agents’ activities. Our approach produces rich, privacy-preserving sensor data that reflects real-world diversity. We evaluate AgentSense on five real HAR datasets. Models pretrained on the generated data consistently outperform baselines, especially in low-resource settings. Furthermore, combining the generated virtual sensor data with a small amount of real data achieves performance comparable to training on full real-world datasets. These results highlight the potential of using LLM-guided embodied agents for scalable and cost-effective sensor data generation in HAR. Our code is publicly available at \url{https://github.com/ZikangLeng/AgentSense}
\end{abstract}

\section{Introduction}

In 1999, \textit{The Matrix} created a simulated reality, one in which most humans lived unknowingly inside a computer-generated illusion. That world was created by intelligent machines, built not to serve humanity, but to control it. Today, we are building simulated environments of our own \footnote{Visualized at \url{https://drive.google.com/file/d/1uab6-Y03uDGwqJ1GV9ooQTIlEAW39pJz/view?usp=sharing}}. But this time, the purpose is different: to understand, model, and support human life through intelligent systems. With the advent of large language models (LLMs) and rich, interactive simulations, we now have the tools to do so.

In 2018, \citet{xia2018gibson} introduced the Gibson Environment, a realistic 3D simulation platform designed for training and evaluating embodied agents—AI systems that perceive and act within physical or simulated environments. Since then, virtual embodied agents (VEAs) have become integral to a wide range of interactive and conversational tasks. These agents take on diverse forms, from virtual 2D or 3D avatars to physical robotic androids equipped with synthetic skin, expressive facial features, and motorized mechanisms for controlling facial expressions and lip movements~\cite{fung2025embodied}. Recent advances have also explored the use of world model-based approaches, which allow agents to ground vision-language prompts within embodied domains and learn complex behaviors through imaginative simulation~\cite{mazzaglia2024genrl}.

Building on the foundation of prior work in embodied AI, we shift the focus from simulating robotic behaviors or dialogue agents to simulating human lives within virtual smart home environments. Our aim is to translate the simulated lives into sensor-level data. Specifically, we focus on smart home based Human Activity Recognition (HAR), which rely on ambient sensors to monitor daily activities. HAR is crucial in domains such as healthcare, elder care, and assisted living~\cite{alam2012review, qi2018examining, chernova2024ai}. However, the development of effective HAR models is limited by the scarcity of large, diverse, and annotated sensor datasets~\cite{liciotti2020sequential, bouchabou2021survey}.

To address this challenge, we simulate diverse human personas and embed them as agents within virtual smart home environments. Each persona is generated using large language models (LLMs), which also direct the agent’s behavior by producing realistic daily routines and corresponding actions grounded in the simulated home environment. These actions are executed within VirtualHome, a 3D simulation platform that we extend with ambient sensor capabilities—referred to as X-VirtualHome.

Each interaction with the environment—opening a door, turning on a light, walking between rooms—triggers corresponding virtual sensor signals, mimicking real-world smart home sensor data. This process allows us to generate large-scale, fully annotated, privacy-preserving datasets without the need for intrusive real-world data collection.

Crucially, our simulation is not solely aimed at generating large volumes of data, but at producing diverse and structured sensor data that capture the heterogeneity of human behavior. By systematically varying personas, daily routines, home layouts, and sensor configurations, our approach generates datasets designed to help HAR models generalize across a wide range of real-world scenarios.

In our experiments, we simulate 18 distinct personas across 22 unique home layouts, generating a total of 250 days of virtual sensor data. We evaluate the resulting models on five real-world smart home datasets—Aruba, Milan, Kyoto7, and Cairo from the CASAS collection~\cite{cook2012casas}, as well as Orange4Home~\cite{cumin2017dataset}—and find that HAR models pretrained on our simulated data consistently outperform baseline methods. Notably, even when fine-tuned with only a small amount of real data, these models achieve performance comparable to those trained entirely on full real-world datasets. These results underscore the practical value of our approach for developing more efficient and scalable HAR systems.

To summarize, through our LLM-based agents in simulated home environments, we make the following contributions:

\begin{enumerate}
\item \textit{An embodied agent framework for smart home data simulation:} We present AgentSense, where LLM-driven agents enact daily lives in virtual smart homes to generate structured, privacy-preserving ambient sensor datasets.
\item \textit{Comprehensive evaluation of virtual sensor data for HAR:} We demonstrate the effectiveness of our approach through extensive experiments on five benchmark smart home datasets. Our results show that models pretrained on virtual data significantly improve HAR performance—especially in low-data regimes—and can match full-data baselines when combined with minimal real-world data.
\end{enumerate}

\section{Related Work}
\subsubsection{Embodied Agents and Simulation Platforms}

Embodied agents are AI systems that perceive, reason, and act within physical or simulated environments. This line of research has become foundational to robotic navigation, task planning, and interactive learning. Simulation platforms have been critical to this progress. The Gibson Environment~\cite{xia2018gibson} introduced a photorealistic 3D simulator for training and evaluating embodied agents in rich environments. Platforms such as AI2-THOR~\cite{Kolve2017thor}, Habitat~\cite{savva2019habitat}, and VirtualHome~\cite{puig_virtualhome_2018} further expanded agent capabilities, enabling object interactions, scripted activities, and complex navigation.

Recent work has broadened the scope of embodied agents beyond physical interaction. Agents now include expressive humanoid avatars and physically embodied androids capable of gaze, speech, and facial expression~\cite{fung2025embodied}. Others integrate world models that support abstract reasoning and future-state simulation~\cite{mazzaglia2024genrl}, improving planning and grounded decision-making.

Despite this progress, simulating human daily routines in smart home contexts for data generation remains underexplored. We build on this direction by generating structured, ambient sensor data from agents enacting daily lives in simulated homes. Rather than creating one-to-one digital twins~\cite{grieves2016digital}, we adopt the digital cousin approach~\cite{dai2024acdc}, simulating diverse agents and environments to support diverse data generation.

A natural extension of this approach is to integrate LLMs to guide agent behavior.

\subsubsection{Language Models for Behavior Simulation}

Recent advances in LLMs have enabled their integration into agent-based systems for reasoning, planning, and interaction. Several works have explored using LLMs to control autonomous agents in virtual environments. For instance, Voyager~\cite{wang_voyager_2023} uses GPT-4 to explore, plan, and act in Minecraft by generating code and updating a skill library. Generative Agents~\cite{park_generative_2023} simulate human behaviors in a virtual town by assigning LLM-driven agents memories, goals, and interactions. CAMEL~\cite{li2023camel} employs role-playing to facilitate multi-agent cooperation toward task completion.

These efforts demonstrate the potential of LLMs to produce structured, plausible agent behaviors. However, most focus on narrative, dialogue, or open-ended exploration rather than generating structured data for downstream tasks, though recent work has examined LLM–human equivalence in annotation behavior~\cite{he2025can}. Our work extends this line by using LLM-guided agents to simulate the daily lives of diverse synthetic personas in virtual smart homes, enabling ambient sensor data generation for training HAR models—a new application of LLM-embodied simulation.

\subsubsection{Synthetic Data Generation for Human Activity Recognition}
Smart home-based Human Activity Recognition (HAR) systems rely on ambient sensors to passively and privacy-preservingly monitor daily activities~\cite{cook2012casas, cumin2017dataset}. However, building robust HAR models is challenging due to the scarcity of large, diverse, annotated datasets that capture variations in home layouts, sensor setups, and resident routines~\cite{liciotti2020sequential, bouchabou2021survey}.

To address data scarcity, recent work has explored synthetic data generation, particularly for wearable HAR—e.g., generating IMU data from video~\cite{kwon2020imutube}, audio~\cite{liang2022audioimu}, and text~\cite{leng2023generating, leng2024imugpt}. However, these methods do not naturally extend to ambient sensors, which involve distinct spatial and triggering mechanisms.

Simulation environments like VirtualHome~\cite{puig_virtualhome_2018} have been used to model household activities. Some approaches generate routines from program sketches~\cite{liao_synthesizing_2019} or use LLMs for daily schedule generation~\cite{yonekura_generating_2024} and action planning~\cite{huang_language_2022}, but none produce ambient sensor data.

Our work bridges this gap by generating synthetic ambient sensor data from LLM-guided agents acting out daily routines in simulated homes. By varying personas, routines, and environments, we produce structured, diverse datasets that better reflect real-world variability.

\section{Methodology}

\begin{figure*}[t]
    \centering
        \includegraphics[width=0.9
        \linewidth]{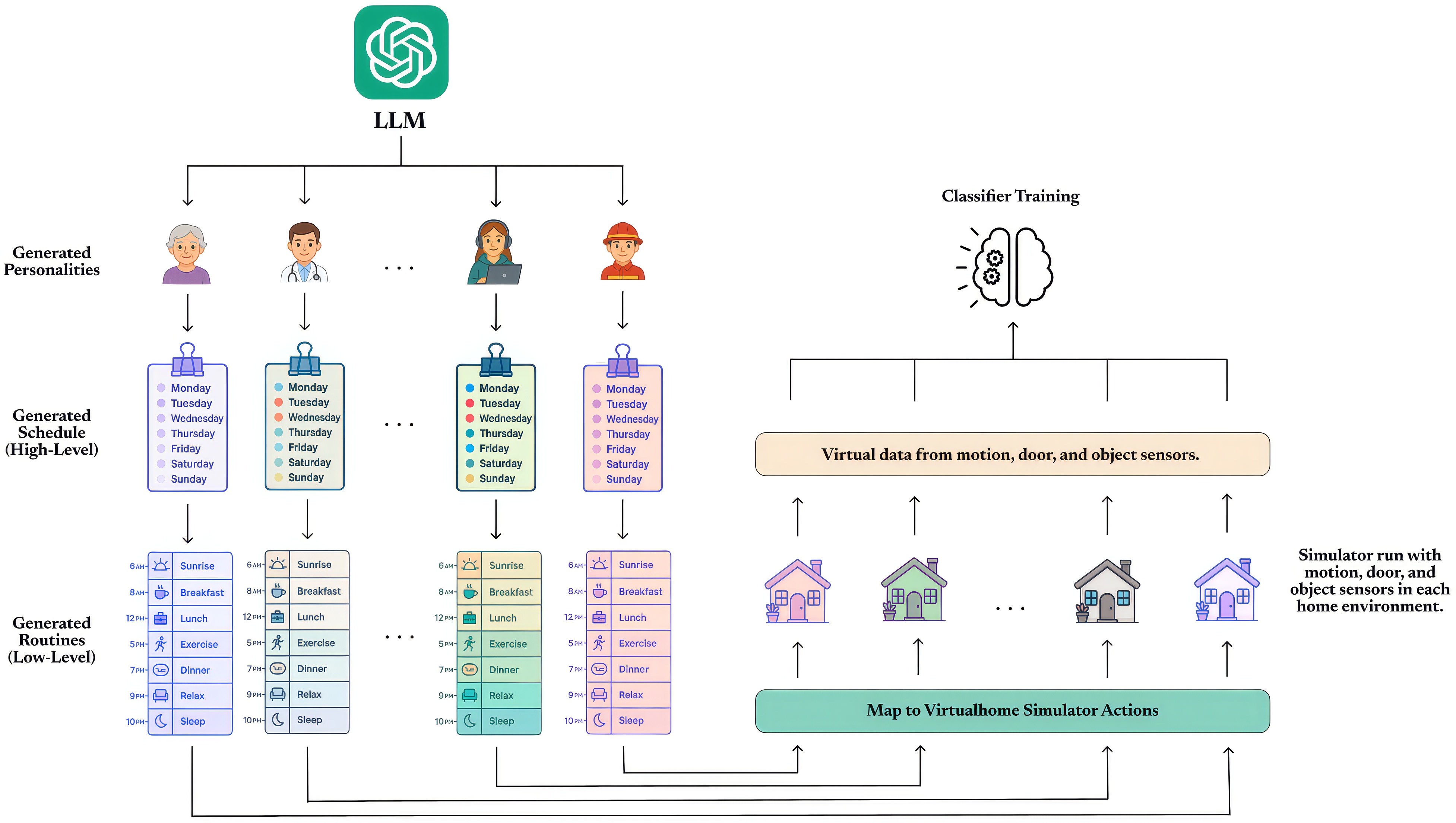}
    \caption{Overview of the framework. The LLM first generates diverse synthetic personas. For each persona, it then produces daily routines grounded in the context of a simulated environment. These routines are decomposed into fine-grained actions, which are executed in the X-VirtualHome simulator. The simulator, augmented with ambient sensors, captures virtual sensor data as the agent enacts its daily life.
    }
    \vspace{-0.1in}
    \label{fig:pipeline}
\end{figure*}

We present a system that uses LLMs and an extended version of the VirtualHome simulator—\textbf{\textit{X-VirtualHome}}—to generate virtual ambient sensor data across diverse home environments and resident profiles (Figure~\ref{fig:pipeline}). Our pipeline prompts LLMs to create personas and daily routines, which are decomposed into simulator-executable action sequences. We extend VirtualHome by adding virtual motion, appliance door, and device activation sensors, enabling the simulation of privacy-preserving ambient data for training activity recognition models.

\subsection{Multi-Level Prompting of LLM Agents to Generate Daily Routines in Diverse Home Environments}
\label{sec: LLM Prompt}

We use a three-stage prompting pipeline to generate diverse, simulator-executable daily routines from LLMs. Starting from generated personas, we produce a high-level schedule, which is then decomposed into fine-grained action sequences compatible with X-VirtualHome.

\subsubsection{Persona Generation }
\label{section:personality_generation}

We aim to capture behavioral diversity crucial for robust HAR models by generating a wide range of personality profiles. Real-world routines vary significantly by age, occupation, health, and lifestyle. Collecting such varied real-user data at scale is costly and time-intensive. Instead, we leverage LLMs to generate realistic personality descriptions, inspired by prior work on persona generation \cite{abbasiantaeb_let_2024, Smrke2025persona, serapio-garcia_personality_2023}. 

\subsubsection{High-Level Daily Routine Generation}
\label{sec:high_level_generation}

We focus on simulating routines for a single resident. To generate a persona-specific daily schedule, we prompt the LLM with:

\begin{enumerate}
    \item \textbf{Persona:} A generated profile including age, occupation, health status, and lifestyle. These attributes shape behavior and introduce meaningful variability for HAR modeling.

    \item \textbf{Day of the Week:} Daily activities differ across the week (e.g., workdays vs. weekends). Specifying the day guides the model to generate contextually appropriate routines.

    \item \textbf{Environment:} A list of rooms in the selected VirtualHome layout. To ensure physical plausibility, the LLM must know what rooms exist. Activities are tagged as ``at home'' or ``outside,'' and only in-home activities are retained for simulation.

    \item \textbf{Example Schedules:} We provide few-shot examples adapted from the Homer dataset \cite{patel2022proactive}, which reflects real human routines. We also instruct the LLM to avoid overly neat time slots (e.g., always ending in 0 or 5) to better mimic natural scheduling.
\end{enumerate}

\subsubsection{Decomposing High-Level Routines into Low-Level Actions}
\label{sec:low_level_generation}

We decompose each high-level activity into a sequence of simulator-executable actions. To do so, we prompt the LLM with the following inputs:

\begin{enumerate}
    \item \textbf{Persona:} Generated personality profile, which includes age, occupation, lifestyle, and health status. These attributes affect how actions are realistically performed—e.g., a retiree may move slower or take more steps.

    \item \textbf{Activity:} A scheduled high-level activity and its start/end time. Only in-home activities are decomposed for compatibility with X-VirtualHome.

    \item \textbf{Environment:} The target room and a list of objects available within it. A separate LLM identifies the appropriate room, and we pass the full object list to ground the model's outputs in the actual environment.

    \item \textbf{Actions:} A predefined set of 18 allowed simulator actions with descriptions (e.g., \texttt{[walk] <bedroom>}) to constrain output format and ensure simulator compatibility.

    \item \textbf{Example Decompositions:} Reference examples demonstrate how to convert high-level tasks into action sequences. These clarify structure and formatting, improving consistency across generated outputs.
\end{enumerate}

\subsection{Converting LLM Output to Executable Actions in VirtualHome}
\label{sec: LLM conversion}

To ensure LLM-generated routines are executable in X-VirtualHome, we convert them into simulator-compatible commands. Left unchecked, raw LLM output may include hallucinated or out-of-vocabulary tokens that cause execution failures. We address this through a five-step process that cleans and validates the output against VirtualHome’s ontology.

\begin{enumerate}
  \item \textbf{Output Cleaning.}
    We remove extraneous metadata (e.g., day labels, high-level activities), retaining only low-level actions annotated with location and timestamp (e.g., \texttt{[walk] <doorjamb> (06:42–06:42) (bedroom)}).

  \item \textbf{Embedding the VirtualHome Vocabulary.} 
    Define
    \[
      \mathcal{A} = \{\text{valid actions}\}, 
      \quad 
      \mathcal{O} = \{\text{valid objects}\}.
    \]
    Each \(x \in \mathcal{A} \cup \mathcal{O}\) is embedded using OpenAI’s \texttt{text-embedding-3-small}:
    \[
      \mathbf{e}(x) = \mathrm{Embed}(x) \in \mathbb{R}^d.
    \]
    These are stored in FAISS indices—one for actions, one for objects \cite{douze2024faiss}.

  \item \textbf{Nearest-Neighbor Retrieval.} 
    For each LLM token \(g\), we compute \(\mathbf{e}(g)\) and retrieve
    \[
      x^* = \arg\max_{x \in \mathcal{V}} \cos(\mathbf{e}(g), \mathbf{e}(x)),
    \]
    where \(\mathcal{V} = \mathcal{A}\) for actions or \(\mathcal{V} = \mathcal{O}_r \subset \mathcal{O}\) for room-specific objects, and
    \[
      \cos(\mathbf{u}, \mathbf{v}) = \frac{\mathbf{u} \cdot \mathbf{v}}{\|\mathbf{u}\| \|\mathbf{v}\|}.
    \]

  \item \textbf{Thresholding and Replacement.}
    Accept \(x^*\) only if 
    \[
      \cos(\mathbf{e}(g), \mathbf{e}(x^*)) \geq \tau,
    \]
    where \(\tau_{\text{act}} = 0.8\) for actions and \(\tau_{\text{obj}} = 0.6\) for objects. Otherwise, the line is flagged as invalid and regenerated by the LLM using surrounding context. After a fixed number of retries, unfixable lines are discarded.

  \item \textbf{Final Command Assembly.}
    Validated tokens replace the originals, yielding simulator-ready commands:
    \[
      \texttt{[action*] <object*> (start--end) (room)}.
    \]
\end{enumerate}

We implement this embedding and grounding workflow using LangChain \cite{langchain} with FAISS and OpenAI embeddings. This ensures semantic alignment with the simulator's ontology, eliminating hallucinations and preserving execution correctness.

\subsection{Implementing Virtual Ambient Sensors in X-VirtualHome Simulator}
\label{sec: X-Virtualhome}

We extend the VirtualHome simulator \cite{puig_virtualhome_2018} to support ambient sensors—specifically motion, appliance door, and device activation sensors—aligned with LLM-generated routines. While VirtualHome allows for the simulation of scripted household activities  and supports multimodal outputs (e.g., action logs, video, segmentation), it lacks support for privacy-preserving sensor simulation. To address this, we integrate virtual motion sensors that track character movement and simulate door/device activations by monitoring environment graph state changes \cite{puig_watch-and-help_2021}.

\subsubsection{Incorporating Motion Sensors in X-VirtualHome}
\label{sec: motion sensor}

We augment each home environment by placing motion sensors based on room size. During simulation, character trajectories are tracked at fixed intervals, and motion detection events are recorded when the character enters a sensor’s detection radius. Below is the detailed procedure:

\begin{enumerate}
    \item \textbf{FindAllRooms}: All rooms in a home environment are detected using VirtualHome’s \texttt{FindAllRooms}, which traverses the hierarchy to find active game objects tagged as \texttt{TYPE\_ROOM}.

    \item \textbf{DetermineSensorCount}: For each room, the area is computed from its bounding box. Sensor count is assigned as follows:
    \begin{itemize}
        \item Small ($\leq 30\,m^2$): 1 sensor
        \item Medium ($30 < \text{area} \leq 60\,m^2$): 2 sensors
        \item Large ($> 60\,m^2$): 3 sensors
    \end{itemize}

    \item \textbf{CalculateSensorPositions}: Sensors are placed near corners of the room, offset by $0.3$\,m horizontally and vertically (from floor and ceiling). Placement varies depending on the number of sensors required.

    \item \textbf{CreateVirtualSensor}: Each sensor is instantiated with a unique ID, room name, and detection radius ($r=5.0$\,m). A \texttt{Motion\_sensor} component is attached, and the sensor is registered in \texttt{MotionSensorManager}.

    \item \textbf{TrackCharacterPosition}: The character’s position $\mathbf{p}(t)$ is logged every $\Delta t = 0.2$\,s using Unity's \texttt{position} property. Motion detection is computed as:
    \[
    \|\mathbf{p}(t) - \mathbf{s}_i\| \leq r
    \]
    where $\mathbf{s}_i$ is the $i$-th sensor's 3D position.

    \item \textbf{MotionDetectionEvent}: On detection, we log: \{\textit{frame number}, \textit{character ID}, \textit{sensor ID}, \textit{room name}, $(x, y, z)$\}. All events are saved for analysis.

    \item \textbf{MotionTriggers}: Using recorded positions, we derive \texttt{ON}/\texttt{OFF} states. A segment $[t_{\text{on}}, t_{\text{off}}]$ is triggered if:
    \[
    \|\mathbf{p}(t) - \mathbf{s}_i\| \leq r \quad \text{and} \quad \|\mathbf{p}(t) - \mathbf{p}(t - \Delta t)\| > \epsilon
    \]
    with $\epsilon = 0.1$\,m for distinguishing motion from jitter. These states are exported as virtual motion sensor readings.
\end{enumerate}

\subsubsection{Simulating Door and Device Activation Sensors}
\paragraph{Environment Graph.}
The Environment Graph is a structured representation of the VirtualHome simulation environment. It encodes objects as nodes and their spatial or semantic relationships as edges. Formally, it is defined as a graph \( G = (V, E) \), where:
\begin{itemize}
    \item \( V \) is the set of nodes, each representing an object instance (e.g., \textit{chair}, \textit{table}, \textit{toothbrush}).
    \item \( E \) is the set of directed edges representing object relations (e.g., \textit{on top of}, \textit{next to}). For example, the object \textit{cup} may be ``ON'' the \textit{kitchen counter}.
\end{itemize}

Each node maintains attributes including the object’s class, associated room, and dynamic states (e.g., \texttt{OPEN}, \texttt{ON}), as well as static properties like \texttt{CAN\_OPEN} and \texttt{HAS\_SWITCH}.

We simulate two types of ambient sensors—door sensors and device activation sensors—by monitoring environment state transitions as recorded by the Environment Graph after each simulated action. Specifically:
\begin{itemize}
    \item A door sensor event is triggered when an object with the property \texttt{CAN\_OPEN} (e.g., doors, cabinets) transitions from \texttt{CLOSED} to \texttt{OPEN}.
    \item A device activation sensor event is triggered when an object with the property \texttt{HAS\_SWITCH} (e.g., microwave, washing machine) transitions from \texttt{OFF} to \texttt{ON}.
\end{itemize}

These virtual sensor events are logged along with metadata such as \textit{timestamp}, \textit{object ID}, \textit{room location}, and updated \textit{object state}. This enables a temporally aligned ambient sensor stream suitable for privacy-preserving HAR model training.

\section{Experiments}
\label{Sec4-Experiments}

To evaluate the effectiveness of the virtual sensor data generated using our approach, we conduct experiments using established real-world smart home datasets. These benchmarks provide a controlled, reproducible setting for assessing model performance and allow us to demonstrate the practical utility of our virtual data generation—without deploying physical hardware. The following sections detail the datasets and classifier training setup.

\subsection{Datasets}

\begin{table*}[t]
  \centering
  \caption{Model performance (Accuracy, Weighted F1, Macro F1) comparing training on real data only versus pretraining on virtual data followed by finetuning on real data, using two TDOST embedding variants.}
  \small
  \begin{tabular}{lccccc}
    \toprule
    & Aruba & Milan & Cairo & Kyoto7 & Orange \\
    \midrule
    \multicolumn{6}{c}{\textbf{Accuracy}} \\
    \midrule
    Real (TDOST‑Basic) & 91.00 $\pm$ 0.53 & 90.07 $\pm$ 0.70 & 69.01 $\pm$ 2.16 & 70.31 $\pm$ 1.53 & 82.40 $\pm$ 0.64 \\
    Real+Virtual (TDOST‑Basic) & \textbf{93.19 $\pm$ 0.22} & \textbf{91.97 $\pm$ 0.42} & \textbf{75.61 $\pm$ 1.93} & \textbf{70.31 $\pm$ 1.53} & \textbf{85.21 $\pm$ 0.82} \\
    
    \midrule
    Real (TDOST‑Temporal) & 91.24 $\pm$ 0.41 & 86.58 $\pm$ 0.12 & 57.20 $\pm$ 1.00 & 48.09 $\pm$ 0.49 & 67.40 $\pm$ 0.39 \\
    Real+Virtual (TDOST‑Temporal) & \textbf{93.67 $\pm$ 0.04} & \textbf{91.60 $\pm$ 0.70} & \textbf{67.10 $\pm$ 2.60} & \textbf{49.31 $\pm$ 0.49} & \textbf{67.60 $\pm$ 0.15} \\

    \midrule
    \multicolumn{6}{c}{\textbf{Macro F1 Score}} \\
    \midrule
    Real (TDOST‑Basic) & 63.98 $\pm$ 0.66 & 70.81 $\pm$ 1.94 & 51.51 $\pm$ 1.58 & 52.48 $\pm$ 1.59 & 21.56 $\pm$ 3.75 \\
    Real+Virtual (TDOST‑Basic) & \textbf{72.20 $\pm$ 0.62} & \textbf{74.44 $\pm$ 0.85} & \textbf{62.47 $\pm$ 1.95} & 
    \textbf{56.07 $\pm$ 1.43} & \textbf{41.83 $\pm$ 2.58} \\
    \midrule
    
    Real (TDOST‑Temporal) & 68.57 $\pm$ 1.21 & 57.20 $\pm$ 1.65 & 21.07 $\pm$ 2.28 & 29.51 $\pm$ 1.99 & 8.42 $\pm$ 2.58 \\
    Real+Virtual (TDOST‑Temporal) & \textbf{77.36 $\pm$ 0.34} & \textbf{73.41 $\pm$ 0.93} & \textbf{46.49 $\pm$ 2.89} & \textbf{31.62 $\pm$ 2.16} & \textbf{10.25 $\pm$ 1.46} \\

    \midrule
    \multicolumn{6}{c}{\textbf{Weighted F1 Score}} \\
    \midrule
    Real (TDOST‑Basic) & 89.81 $\pm$ 0.55 & 90.20 $\pm$ 0.69 & 66.79 $\pm$ 1.83 & 66.43 $\pm$ 1.48 & 75.91 $\pm$ 0.86 \\
    Real+Virtual (TDOST‑Basic) & \textbf{92.41 $\pm$ 0.18} & \textbf{91.74 $\pm$ 0.30} & \textbf{74.65 $\pm$ 1.98} & \textbf{68.27 $\pm$ 1.02} & \textbf{83.42 $\pm$ 0.76} \\

    \midrule
    Real (TDOST‑Temporal) & 89.57 $\pm$ 0.40 & 84.99 $\pm$ 0.33 & 44.45 $\pm$ 0.29 & 41.70 $\pm$ 1.46 & 61.68 $\pm$ 5.73 \\
    Real+Virtual (TDOST‑Temporal) & \textbf{93.48 $\pm$ 0.12} & \textbf{91.23 $\pm$ 0.64} & \textbf{61.14 $\pm$ 2.85} & \textbf{43.96 $\pm$ 1.68} & \textbf{65.68 $\pm$ 0.43} \\
    \bottomrule
  \end{tabular}
  \label{tab:model_perf}
\end{table*}

\subsubsection{Real Datasets}
We conduct experiments on five publicly available datasets: Aruba, Milan, Kyoto7, and Cairo from the CASAS collection~\cite{cook2012casas}, and Orange4Home~\cite{cumin2017dataset} from the Amiqual4Home environment. Among the CASAS datasets, Aruba has the most data points and balanced sensor modalities (motion, door, temperature), with a floorplan similar to Milan, which shares the same sensor types. Kyoto7, Cairo, and Orange are multi-story homes with more diverse layouts and fewer samples. Cairo and Kyoto7 include two residents, adding behavioral variability, while Orange has one. Sensor types also vary: Cairo includes motion and temperature; Kyoto7 adds item usage, light switches, and device activations; Orange features 18+ modalities, including noise, voltage, and humidity.

\subsubsection{Virtual Dataset}
We generated virtual sensor data for 18 distinct personas across 22 simulated home environments, yielding 250 days of activity data. The final dataset includes 3,266 activity windows, each containing between 3 and 393 sensor triggers (average: 36). All environments represent single-story homes with a single resident.

During data generation, LLM agents freely produced open-ended routines without restrictions on the activity space. To align these with real-world benchmarks, we mapped the LLM-generated activities to the label sets defined in the target HAR datasets. This was done by prompting the LLM with (1) the high-level activity name, (2) its decomposed sequence of low-level actions, and (3) the complete set of activity labels from the target dataset. The LLM then selected the most appropriate label, which we assigned to the corresponding virtual sensor data.

Since the simulation assumes single-resident settings, overlapping activities from multi-resident datasets were interpreted as being performed by one individual. Additionally, when real-world activity labels were not reflected in the virtual data—due to unconstrained routine generation—we assigned them the label “Other.”

\subsection{Classifier Training:} 

We adopt the TDOST-based HAR framework proposed by \citet{thukral2025layout}, which transfers across diverse home layouts and sensor configurations. Unlike transformer- or graph-based models that assume fixed topologies, TDOST is layout-agnostic, making it well-suited for cross-environment evaluation. This allows us to directly assess the impact of pretraining on virtual data.

We use a pre-segmented, activity-level windowing approach, selecting the first 100 sensor triggers from each activity window. Each trigger includes contextual metadata (e.g., sensor type, location, and timestamp), which we convert into natural language sentences using two TDOST variants:
\begin{itemize}
    \item \textbf{TDOST-Basic:} Encodes sensor type and location. For example, \textit{"Motion sensor in bedroom fired with value ON"}.
    \item \textbf{TDOST-Temporal:} Adds time information to the above, e.g., \textit{"Motion sensor in bedroom fired with value ON at twelve hours six minutes PM"}.
\end{itemize}

These sentences are embedded using the \texttt{all-distilroberta-v1} model from SentenceTransformers~\cite{reimers2019sentencebertsentenceembeddingsusing}. The resulting sequence of embeddings is passed into a bidirectional LSTM (Bi-LSTM) with 64 hidden units, following the architecture used by \citet{thukral2025layout}. Finally, a linear classification layer maps the encoded sequence to a probability distribution over activity classes.

\subsubsection{Training Settings}
To evaluate the effectiveness of the generated virtual sensor data, we conduct two types of experiments across all real-world datasets: \textit{Real} and \textit{Real + Virtual}. In the \textit{Real} setting, the model is trained and evaluated in a fully supervised manner on each dataset independently, serving as our baseline. In the \textit{Real + Virtual} setting, we follow a two-stage training procedure inspired by \citet{kwon2020imutube}: the model is first pretrained on virtual sensor data, then fine-tuned on real sensor data, with all weights updated. Final evaluation is performed on the real test split.

\subsection{Results}

From Table \ref{tab:model_perf}, we note that models pretrained with virtual data, Real + Virtual (TDOST-Basic) and Real + Virtual (TDOST-Temporal), consistently outperform their counterparts trained exclusively on real data across the  benchmark HAR datasets.

For TDOST-Basic, the average accuracy increases across datasets, with notable gains such as 69.01\% to 75.61\% on Cairo, and 82.40\% to 85.21\% on Orange4Home.
Similarly, TDOST-Temporal shows substantial improvements, especially on low-resource datasets, as such accuracy improves by 10\%  on Cairo and 5\% on Milan. while performance on other datasets such as  Kyoto7 and Orange4Home remains stable.
Macro F1 Score shows substantial gains using our Virtual+Real supervised  pipeline. 
For TDOST-Basic, the Macro F1 increases from 11\% on Cairo and by approximately ~20\% on the Orange4Home dataset.
Similarly, TDOST-Temporal shows a sharp improvement from 68.57\% to 77.36\% for Aruba, 51\% to 73\% for Milan. 
Weighted F1 scores follow a similar upward trend, with both TDOST-Basic and TDOST-Temporal showing significant improvements across datasets.

The consistency of improvements across all five datasets—including Orange4Home, which is a large-scale non-CASAS dataset with over 18 different types of sensors—further reinforces the generalizability of pretraining and robustness of our virtual sensor data across different sensor types, activity sets, and home environments.
To sum up, substantial improvements in downstream HAR performance validate our virtual data-generation approach as an effective way to reduce reliance on costly real-world data collection, especially for complex or low-resource environments.

\subsection{Ablation Studies}

\subsubsection{ Varying the Amount of Real Data Used for Finetuning}
\label{sec:vary real}

\begin{figure*}[t]
    \centering
        \includegraphics[width=0.9\linewidth]{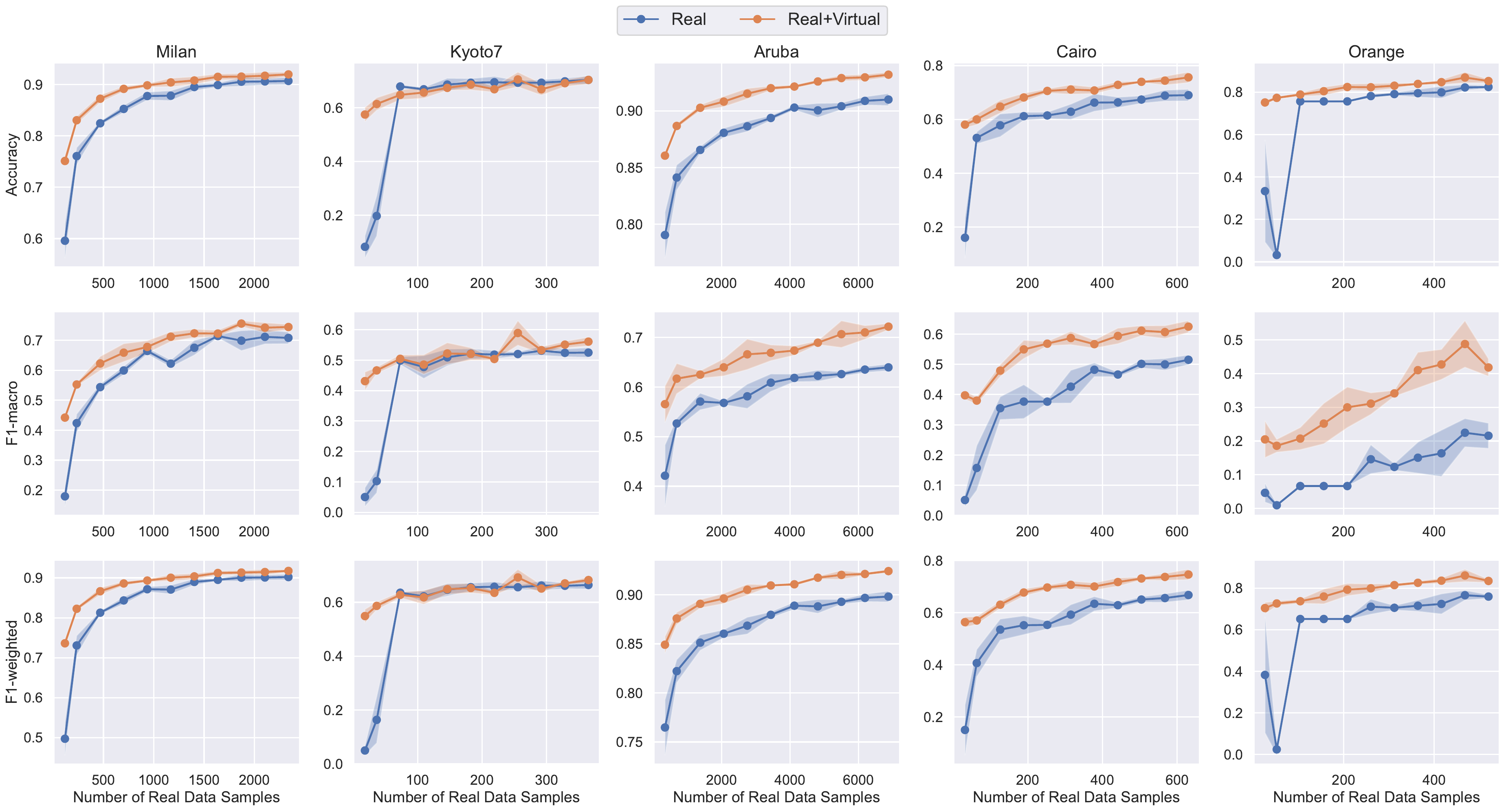}
    \caption{TDOST Basic model performance when different amount of real data are used for training. The amount of virtual data stays the same.  }
    \vspace{-0.2in}
    
    \label{fig:basic_vary_real}
\end{figure*}

We analyze how varying the amount of real data used during fine-tuning affects downstream HAR performance. This experiment aims to identify the minimum quantity of real-world data needed to achieve competitive performance when models are pretrained on virtual data.

For each dataset, we first pretrain the model using all available virtual data, then fine-tune it using randomly sampled subsets of real data. The number of fine-tuning samples varies per dataset, as illustrated in Figure~\ref{fig:basic_vary_real}. To ensure consistency, the sampled subsets maintain the original distribution of activity classes. All experiments are conducted using the TDOST-Basic variant.

Across all five benchmark datasets, we observe that the \textit{Real+Virtual} models consistently outperform those trained solely on real data, regardless of the amount used. This indicates that pretraining on virtual data provides a strong initialization, enabling robust performance even in low-data regimes. Notably, using only $5\%$ to $10\%$ of real data, pretraining on virtual data yields substantial gains over training using only the real data. For example, we see a $\sim$10\% improvement in macro F1 for Aruba and up to a 45\% increase for Kyoto7 when using just $5\%$ of real data. These findings support our hypothesis that virtual data can significantly reduce the need for large-scale real-world data collection.

In several cases, such as Orange4Home and Cairo, \textit{Real+Virtual} performance approaches the upper bound of fully supervised training. On Cairo, for instance, models reach $\sim$80\% accuracy and $\sim$60\% macro F1 using just 200 real samples. Similarly, fewer than 200 samples are sufficient to achieve comparable results on Orange. This further underscores the practical value of virtual pretraining for efficient HAR model development.

\begin{table}[ht]
\centering
\small
\begin{tabular}{ccc|ccc}
\toprule
   Env & Days &  Personas & Acc & Macro F1 & Weighted F1\\
\midrule
 & & & 92.72 & 68.35 & 91.66 \\
\checkmark &  & & 92.76 & 70.69 & 91.84 \\
\checkmark &\checkmark &  &  93.22 & 71.01 & 92.38 \\
\checkmark & \checkmark & \checkmark & 93.19 & 72.20 & 92.41 \\
\bottomrule
\end{tabular}
\caption{Effect of environment diversity, weekly routine coverage, and persona variation on HAR performance}
\vspace{-0.1in}
\label{tab:ablation}
\end{table}

\subsubsection{Effectiveness of Individual Components} 
We evaluate the impact of different generation settings—namely, personas, routines from multiple days of the week, and diverse home environments—on the effectiveness of virtual sensor data. All experiments are conducted on the Aruba dataset using the TDOST-Basic variant. As shown in Table~\ref{tab:ablation}, when virtual data is generated using a single persona, a single day’s routine, and a single environment, the downstream classifier achieves a macro F1 score of 68.35\%. Introducing diversity in environments (22 homes) increases performance to 70.69\%. Adding routines from all seven days of the week further improves macro F1 to 71.01\%. Finally, incorporating multiple personas leads to the highest performance of 72.20\% macro F1.

These results indicate that each component—environment, daily routine, and persona—contributes positively to the performance of the downstream classifier. We attribute this improvement to increased diversity in the generated virtual sensor data, which enhances the generalizability of the downstream classifier. Notably, the total volume of virtual sensor data remains constant across settings, achieved through repeated generation.

\section{Conclusion}

We introduced \textit{AgentSense}, a data generation framework that leverages LLM-guided embodied agents to simulate human lives in virtual smart home environments. By generating diverse synthetic personas and their daily routines grounded in varied home layouts, \textit{AgentSense} produces structured, richly annotated ambient sensor data designed to reflect the heterogeneity of real-world settings. We extended the VirtualHome simulator with virtual ambient sensors, enabling the conversion of LLM-generated action sequences into sensor data.

Our experiments across five real-world smart home datasets demonstrate that models pretrained on this virtual data consistently improve downstream HAR performance—particularly in low-resource scenarios. Even with limited real data, these models approach the performance of those trained on full datasets, highlighting the practical value of our approach. This work illustrates a promising new direction for scalable, privacy-preserving HAR development: using LLM-guided embodied simulation to reduce dependence on costly real-world data collection. 

\section*{Acknowledgments}
This work was partially supported by NSF IIS-2112633 and the NSF Research Fellowship under Grant No. DGE-2039655. 

Any opinion, findings, and conclusions or recommendations expressed in this material are those of the authors(s) and do not necessarily reflect the views of the National Science Foundation.

\bibliography{./bibs/ref}


\appendix

\clearpage
\clearpage

\section{Full Prompt Template for Personality Generation}
\label{app:personality_prompt}

Below are the exact system and user prompts used for persona generation.



\begin{tcolorbox}[
  title=\centering{\textcolor{black!100}{\texttt{GPT 4o-mini} Persona Generation Prompt}} ,
  colback=brown!5!white,         
  colframe=brown!15!white,  
  coltitle=black,
  fonttitle=\bfseries,
  sharp corners=south,
  listing only,
  listing options={
    basicstyle=\ttfamily\small,
    breaklines=true,
    showstringspaces=false
  }
]
\small
\begin{verbatim}
--SYSTEM-- 
You are a personality generator that 
generates descriptions of people based 
on the specified format. Each 
description should include the person’s 
name, age, job, health status (whether 
physical or mental), and a description 
of their lifestyle, habits, and 
personality traits. The total length 
must be between 85 and 100 words. 
Remember to include proper punctuation,
but do not use asterisks anywhere in 
the sentences!!!

--USER-- 
Each description should strictly follow
this format: 'Person's name, he/she is 
XX years old, job description, and has 
the following health situation: 
[physical or mental conditions, or 
none].' The description should include 
details about the person's lifestyle, 
habits, and personality traits. Make 
sure the age, job, health condition, 
and lifestyle vary between people, and 
the descriptions should be unique and 
diverse.
\end{verbatim}
\end{tcolorbox}

\section{Full Prompt Template for Schedule Generation}
\label{app:schedule_prompt}

Below is the exact prompt sent to the LLM on each day, with placeholders for the variable parts and the complete example routines:

\begin{tcolorbox}[
  title=\centering{\textcolor{black!100}{\texttt{GPT 4o-mini} Schedule Generation Prompt}} ,
  colback=brown!5!white,
  colframe=brown!15!white,
  coltitle=black,
  fonttitle=\bfseries,
  sharp corners=south,
  breakable,
  listing only,
  listing options={
    basicstyle=\ttfamily\small,
    breaklines=true,
    breakatwhitespace=false,
    showstringspaces=false
  }
]
\small
\begin{verbatim}
--SYSTEM--
Generate a detailed daily schedule 
based on a person’s personality, 
home environment, and the specific 
day of the week. The schedule should 
cover the entire day, from wake-up to 
sleep, and reflect a realistic 
lifestyle with variations between 
weekdays and weekends.

--USER--
The personality description is as 
follows: "{personality_description}"
Based on the provided description 
(job, age, health, and personality), 
generate a daily schedule for this 
person on {day}. The schedule should 
cover the entire day, from wake-up 
to sleep, using the 24-hour time 
format. For each time slot, vary the 
start and end times, avoiding times
divisible by 5 or 10 minutes. 
Activities happening at home must be
in one of the provided rooms: 
{room_list}. If an activity does not 
occur in one of these rooms, it must 
happen 'outside.' The character is 
the only person at home, so do not 
generate events involving family or 
friends. Use the following format: 
'Activity Description (Start Time - 
End Time) (at home/outside).' Two 
examples are provided below. Generate 
the schedule in this format only; do 
not include any additional information.

Example 1:
wake_up (06:47 - 06:50) (at home)
brushing_teeth (06:50 - 06:56) 
(at home)
going_to_the_bathroom (06:56 - 07:12) 
(at home)
yoga_practice (07:12 - 07:41) 
(at home)
showering (07:41 - 08:02) (at home)
breakfast (08:02 - 08:30) (at home)
commuting_to_work (08:30 - 09:14) 
(outside)
computer_work (09:14 - 11:45) (outside)
going_to_the_bathroom (11:45 - 11:52) 
(outside)
lunch_with_colleagues (11:52 - 12:48) 
(outside)
catching_up_on_emails (12:48 - 14:06) 
(outside)
team_meeting (14:06 - 15:30) (outside)
brainstorming_session (15:30 - 16:17) 
(outside)
going_to_the_bathroom (16:17 - 16:25) 
(outside)
computer_work (16:25 - 18:00) (outside)
commuting_home (18:00 - 18:42) 
(at home)
going_to_the_bathroom (18:42 - 18:49) 
(at home)
vegan_cooking_experiment 
(18:49 - 20:06) (at home)
cleaning_up_after_cooking 
(20:06 - 20:30) (at home)
playing_computer_games (20:30 - 22:40) 
(at home)
wind_down_time (22:40 - 23:10) (at home)
reading (23:10 - 00:33) (at home)
brushing_teeth (00:33 - 01:14) (at home)
sleep (01:14) (at home)

Example 2:
wake_up (10:01 - 10:07) (at home)
brushing_teeth (10:07 - 10:16) (at home)
showering (10:16 - 10:32) (at home)
commuting_to_class (10:32 - 10:55) 
(outside)
environmental_science_class 
(10:55 - 11:50) (outside)
going_to_the_bathroom (11:50 - 11:54) 
(outside)
studying_in_library (11:54 - 12:30) 
(outside)
lunch (12:30 - 13:08) (outside)
commuting_to_work (13:08 - 16:25) 
(outside)
going_to_the_bathroom (16:25 - 16:31) 
(outside)
finishing_work (16:31 - 16:42) (outside)
commuting_home (16:42 - 17:05) (outside)
going_to_the_bathroom (17:05 - 17:08) 
(at home)
dinner_preparation (17:08 - 17:40) 
(at home)
dinner (17:40 - 18:28) (at home)
socializing_with_friends_via_video_call 
(18:28 - 19:44) (at home)
organizing_photography_ideas 
(19:44 - 20:45) (at home)
musical_exploration_time (20:45 - 21:21) 
(at home)
going_out_for_an_evening_walk 
(21:21 - 22:04) (outside)
wind_down_with_a_book (22:04 - 23:00) 
(at home)
do_homework (23:00 - 03:00) (at home)
brushing_teeth (03:00 - 03:07) (at home)
sleep (03:07) (at home)
\end{verbatim}
\end{tcolorbox}

\section{Activity Breakdown Prompt}
\label{app:decompose_prompt}

Below is the exact system and user prompt used to decompose each high‑level activity into fine‑grained steps.

\begin{tcolorbox}[
  title=\centering{\textcolor{black!100}{\texttt{GPT 4o-mini} Activity Breakdown Prompt}} ,
  colback=brown!5!white,
  colframe=brown!20!white,
  coltitle=black,
  fonttitle=\bfseries,
  sharp corners=south,
  breakable,
  listing only,
  listing options={
    basicstyle=\ttfamily\small,
    breaklines=true,
    breakatwhitespace=false,
    showstringspaces=false
  }
]
\small
\begin{verbatim}
--SYSTEM--
Based on the start time, end time, a 
brief activity description, and the 
specific room where this activity 
takes place (as all activities occur 
at home), your task is to break the 
activity into detailed steps, imagining 
you are the person following the 
schedule. You will also be provided 
with the person’s description, a list 
of available objects in the room, and a 
predefined set of action verbs and 
structures that must be strictly 
followed to ensure consistency and 
accuracy in the breakdown.

--USER--
Now you are provided with the following 
details:

Activity Name: {activity}
Start Time: {start_time}
End Time: {end_time}
Personality Description: 
{personality_description}
Location: {selected_room}

Based on the 'Activity Name,' break 
this activity into detailed action 
steps corresponding to smaller time 
intervals (24-hour format) within the 
'Start Time' and 'End Time'. For each 
smaller time interval, use the 
following format:

First line: 'Start time - End time, 
{selected_room}'

Then, in the following lines, describe 
the activities during that smaller time 
interval. Each action step should also 
have its own time slot. The 
'Personality Description' is provided 
for reference. Each activity step can 
only use objects from this list: 
{objects_str}. Additionally, each step 
must adhere to one of the structural 
formats below. If an action requires 
a verb that is not on the format list, 
replace it with the closest matching 
verb from the list. Under no 
circumstances should you introduce 
new verbs or deviate from the defined 
structures. Below are all the format 
structures:

[walk] <object> 
(e.g., walk to the bedroom should be 
[walk] <bedroom>)

[run] <object> 
(e.g., run to the kitchen should be 
[run] <kitchen>)

[walkforward] 
(just walk forward; no objects should 
follow)

[turnleft] 
(just turn left; no objects should 
follow)

[turnright] 
(just turn right; no objects should 
follow)

[sit] <object> 
(e.g., sit on the chair should be 
[sit] <chair>)

[standup] 
(just stand up; no objects should 
follow)

[grab] <object> 
(e.g., grab the apple should be 
[grab] <apple>)

[open] <object> 
(e.g., open the fridge should be 
[open] <fridge>)

[close] <object> 
(e.g., close the fridge should be 
[close] <fridge>)

[put] <object 1> <object 2> 
(e.g., put the apple on the table 
should be [put] <apple> <table>)

[switchon] <object> 
(e.g., switch on the stove should 
be [switchon] <stove>)

[switchoff] <object> 
(e.g., switch off the stove should 
be [switchoff] <stove>)

[drink] <object> 
(e.g., drink from the waterglass 
should be [drink] <waterglass>)

[touch] <object> 
(e.g., touch the stove should be 
[touch] <stove>)

[lookat] <object> 
(e.g., look at the stove should be 
[lookat] <stove>)

Please follow the structure. Verbs are 
already provided in the format list 
above, and objects can only be selected 
from the object list above.

Here is an example of an activity 
breakdown, where the 'Activity Name' 
is 'brushing_teeth', the 'Start Time' 
is 7:20, and the 'End Time' is 7:26. 
As you can see, the start time of the 
first step in each interval aligns with 
the interval's start time, and the last 
step of each interval ends with the 
interval's end time.

7:20 - 7:22, bathroom
Step 1: [walk] <bathroom> 
(7:20 - 7:20)
Step 2: [switchon] <lightswitch> 
(7:20 - 7:20)
Step 3: [walk] <bathroomcounter> 
(7:20 - 7:21)
Step 4: [grab] <toothbrush> 
(7:21 - 7:21)
Step 5: [lookat] <toothpaste> 
(7:21 - 7:22)

7:22 - 7:26, bathroom
Step 1: [grab] <toothpaste> 
(7:22 - 7:22)
Step 2: [put] <toothpaste> <toothbrush> 
(7:22 - 7:23)
Step 3: [drink] <waterglass> 
(7:23 - 7:25)
Step 4: [put] <waterglass> 
<bathroomcounter> (7:25 -7:25)
Step 5: [switchoff] <lightswitch> 
(7:26 - 7:26)

The format must be exactly like this. 
Do not generate any other sentences!!!!!
\end{verbatim}
\end{tcolorbox}

Below is the exact system and user prompt used to select the most appropriate room for a given activity.

\begin{tcolorbox}[
  title=\centering{\textcolor{black!100}{\texttt{GPT 4o-mini} Room Detection Prompt}} ,
  colback=brown!5!white,
  colframe=brown!20!white,
  coltitle=black,
  fonttitle=\bfseries,
  sharp corners=south,
  breakable,
  listing only,
  listing options={
    basicstyle=\ttfamily\small,
    breaklines=true,
    breakatwhitespace=false,
    showstringspaces=false
  }
]
\small

\begin{verbatim}
--SYSTEM--
You will be provided with a list of 
rooms and an activity name. Your 
task is to determine which room is 
most likely to be the location for 
the given activity.

--USER--
Your task is to determine the most 
appropriate room for a given 
activity from the provided list. 
Below are the activity description 
and the room list:

Activity: {activity}
Room list: {room_list_str}

Return the name of the room that is
most suitable for this activity 
from the list above. You must select 
and return exactly one room name. Do 
not include any explanations or 
additional information, just the room 
name.
\end{verbatim}
\end{tcolorbox}

\section{Full Prompt Template for Label Generation}
\label{app:label_prompt}

Below is the exact prompt sent to the LLM to generate a label on each action block. 

\begin{tcolorbox}[
  title=\centering{\textcolor{black!100}{\texttt{GPT 4o-mini} Label Generation Prompt}} ,
  colback=brown!5!white,
  colframe=brown!15!white,
  coltitle=black,
  fonttitle=\bfseries,
  sharp corners=south,
  breakable,
  listing only,
  listing options={
    basicstyle=\ttfamily\small,
    breaklines=true,
    breakatwhitespace=false,
    showstringspaces=false
  }
]
\small
\begin{verbatim}
--SYSTEM--
You are an intelligent assistant 
helping label smart home activities. 
You will be provided with an 'Activity 
Name' and its corresponding detailed 
routine steps. Your task is to choose 
ONE label from the provided set that 
best describes the activity. Return 
ONLY the label (exactly the same word 
appears in the set). Do NOT provide 
any explanation.

--USER--
Activity Name: {activity_name}
Routine Block: {routine_text}
Label Set: 
{aruba_labels/milan_labels/cairo_labels
/kyoto7_labels/orange_labels}

Note: The label 'bed_to_toilet' refers 
to activities that involve walking from 
the bed to the bathroom.

Which label from the list best fits 
this activity? Please return ONLY the 
label (exactly the same word appears 
in the set). Do NOT provide any 
explanation.

\end{verbatim}
\end{tcolorbox}

Below is an example of our input block and the generated labels:

\begin{tcolorbox}[
  title=\centering{\textcolor{black!100}{\texttt{GPT 4o-mini} Generated Labels for an Input Action Block}} ,
  colback=brown!5!white,
  colframe=brown!15!white,
  coltitle=black,
  fonttitle=\bfseries,
  sharp corners=south,
  breakable,
  listing only,
  listing options={
    basicstyle=\ttfamily\small,
    breaklines=true,
    breakatwhitespace=false,
    showstringspaces=false
  }
]
\small
\begin{verbatim}
Input Action Block:
[walk] <kitchen> (07:10 - 07:10) 
(kitchen)
[switchon] <coffeemaker> 
(07:10 - 07:10) (kitchen)
[standup] (07:10 - 07:11) (kitchen)
[grab] <waterglass> (07:11 - 07:11) 
(kitchen)
[drink] <waterglass> (07:11 - 07:12) 
(kitchen)
[put] <waterglass> <kitchencounter> 
(07:12 - 07:13) (kitchen)
[walk] <fridge> (07:13 - 07:13) 
(kitchen)
[open] <fridge> (07:13 - 07:14) 
(kitchen)
[grab] <bananas> (07:14 - 07:15) 
(kitchen)
[close] <fridge> (07:15 - 07:15) 
(kitchen)
[put] <bananas> <kitchencounter> 
(07:16 - 07:16) (kitchen)
[walk] <toaster> (07:16 - 07:17) 
(kitchen)
[switchon] <toaster> (07:17 - 07:17) 
(kitchen)
[grab] <breadslice> (07:17 - 07:18) 
(kitchen)
[put] <breadslice> <toaster> 
(07:18 - 07:18) (kitchen)
[lookat] <toaster> (07:20 - 07:20) 
(kitchen)
[walk] <kitchentable> (07:20 - 07:21) 
(kitchen)
[sit] <kitchentable> (07:21 - 07:21) 
(kitchen)
[lookat] <coffeemaker> (07:21 - 07:22) 
(kitchen)
[grab] <waterglass> (07:22 - 07:23) 
(kitchen)
[drink] <waterglass> (07:26 - 07:26) 
(kitchen)
[put] <waterglass> <kitchencounter> 
(07:26 - 07:27) (kitchen)
[grab] <coffeepot> (07:27 - 07:30) 
(kitchen)

LLM-Generated Labels:
Aruda: Eating
Cairo: Breakfast
Milan: Kitchen Activity
Kyoto7: Meal Preparation
Orange: Cooking

\end{verbatim}
\end{tcolorbox}

\section{Real Datasets}
\label{app: real datasets}
We specify the details of Smart Home datasets under evaluation in our work. We detail more the type of sensors, floor layouts and activities in Table \ref{tab:datasets}. This table has been taken from \cite{thukral2025layout} with permission.

\begin{table*}[h]
\caption{ This table summarizes the datasets used in our experiments. The sensors $M$, $D$, $T$, $I$, $LS$, $AD$, $S$, $MD$, $N$, $V$, $PR$, $SH$, $L$, $P$, $DR$, $H$, $HE$, $LM$, $C$, $WT$, and $W$ correspond to motion, door, temperature, item, light switch, activate device (burner, hot water, cold water), setpoint, mode, noise, voltage, presence, drawer, humidity, heater, luminosity, cupboard, water, and window sensors, respectively. The datasets differ in sensor types, sensor placement, number of residents, home layouts, and recorded activities. The number of activity datapoints is as in the original dataset. This table has been taken from \cite{thukral2025layout} with permission.}
\centering
\footnotesize
\renewcommand{\arraystretch}{1.2}
\setlength{\tabcolsep}{2pt}
\begin{tabular}{p{1.2cm} p{3cm} p{0.8cm} p{3.5 cm} p{6cm}}
    \toprule
    Dataset & Sensors & Res. & Floorplan & Activities \\
    \midrule
    Aruba & $[M, D, T]$ & 1 & Single story home with living space, dining space, kitchen, office, 2 bedrooms, 2 bathrooms, and closet & Relax (2919), Meal Preparation (1606), Enter Home (431), Leave Home (431), Sleeping (401), Eating (257), Work (171), Bed to Toilet (157), Wash Dishes (65), Housekeeping (33), Resperate (6), Other (6354)\\
    \hline
    Milan & $[M, D, T]$ & 1 & Single story with living space, dining, kitchen, workspace/TV room, 2 bedrooms, 2 bathrooms, and closet & Kitchen Activity (554), Guest Bathroom (330), Read (314), Master Bathroom (306), Leave Home (214), Master Bedroom Activity (117), Watch TV (114), Sleep (96), Bed to Toilet (89), Desk Activity (54), Morning Meds (41), Chores (23), Dining Room Activity (22), Evening Meds (19), Meditate (17), Other (1943)  \\
    \hline
    Kyoto7 & $[M, D, T I, LS, AD]$ & 2 & Double story with living, dining, kitchen, pantry, closet, 2 bedrooms, office, bathroom & Meal Preparation, R1 Work (59), R1 Personal Hygiene (44), R2 Work (44), R2 Bed to Toilet (39), R2 Personal Hygiene (38), R1 Sleep (35), R2 Sleep (35), R1 Bed to Toilet (34), Watch TV (30), Study (9), Clean (2), Wash Bathtub (1) \\
    \hline
    Cairo & $[M, T]$ & 2 & Three story with living, dining, kitchen, 2 bedrooms, office, laundry, garage & Leave Home (69), Night Wandering (67), R1 Wake (53), R2 Wake (52), R2 Sleep (52), R1 Sleep (50), Breakfast (48), R1 Work in Office (46), R2 Take Medicine (44), Dinner (42), Lunch (37), Bed to Toilet (30), Laundry (10) \\
    \hline
    Orange & \begin{tabular}[t]{@{}l@{}}$[S, WT, MD], [D, V, PR],$\\$[P, DR, W], [H, HE, LM],$\\$[LS, T, N], [SH, L, C]$\end{tabular} & 1 & Double story with living room, kitchen, entrance, staircase, toilet, bathroom, walkway, bedroom, and office & 
    \{Bathroom:\} Cleaning (4), Showering (19), Using Sink (38), Using Toilet (9), \{Bedroom:\} Cleaning (3), Dressing (30), Napping (15), Reading (15), \{Entrance:\} Entering (21), Leaving (21), \{Kitchen:\} Cleaning (4), Cooking (19), Preparing (19), Washing Dishes (19), \{Living Room:\} Cleaning (19), Computing (15), Eating (19), Watching TV (18), \{Office:\} Cleaning (4), Computing (46), Watching TV (14), \{Staircase:\} Going Up (57), Going Down (57), \{Toilet:\} Use Toilet (11), Other (478) \\
    \bottomrule
\end{tabular}
\label{tab:datasets}
\end{table*}

\section{Mapping of Activities}
\label{app: Activity Mapping}
The set of activities generated by the LLM may not exactly match those in the real-world datasets (Aruba, Milan, Kyoto7, Cairo, Orange). When mapping, we pair a real-world activity with its VirtualHome counterpart if it was generated; otherwise, we assign it to the “Other” category. All mapping activities for each dataset are in Table \ref{tab:activity_mapping}.

\begin{table*}[h]
\centering
\caption{Mapping of activities for each dataset.}
\label{tab:activity_mapping}
\begin{tabularx}{\textwidth}{X|X|X|X|X}
\toprule
\textbf{Aruba} & \textbf{Milan} & \textbf{Kyoto7} & \textbf{Cairo} & \textbf{Orange} \\
\midrule
Relax → Relax & Kitchen Activity → Kitchen Activity  & Meal Preparation → Meal Preparation & Leave Home → Leave Home & Cleaning → Cleaning \\

Meal Preparation → Meal Preparation & Guest Bathroom → Guest Bathroom  & R1 Work → Work & Night Wandering → Night Wandering & Showering → Showering \\

Enter Home → Other & Read → Read & R1 Personal Hygiene → Personal Hygiene & R1 Wake → Wake & Using Sink → Using Sink \\

Leave Home → Leave Home & Master Bathroom → Master Bathroom & R2 Work → Work & R2 Wake → Wake & Using Toilet → Using Toilet \\

Sleeping → Sleeping & Leave Home → Leave Home & R2 Bed to Toilet → Bed to Toilet & R1 Sleep → Sleep & Dressing → Dressing \\

Eating → Eating & Master Bedroom Activity → Master Bedroom Activity & R2 Personal Hygiene → Personal Hygiene & R2 Sleep → Sleep & Napping → Napping \\

Work → Work & Watch TV → Other & R1 Sleep → Sleep & Breakfast → Breakfast & Reading → Reading \\

Bed to Toilet → Bed to Toilet & Sleep → Sleep & R2 Sleep → Sleep & R1 Work in Office → Work in Office & Entering → Other \\

Wash Dishes → Other & Bed to Toilet → Bed to Toilet & R1 Bed to Toilet → Bed to Toilet & R2 Take Medicine → Take Medicine & Leaving → Leaving \\

Housekeeping → Housekeeping & Desk Activity → Other & Watch TV → Other & Dinner → Dinner & Kitchen Preparing → Kitchen Preparing \\

Resperate → Other & Morning Meds → Other & Study → Study & Lunch → Lunch & Washing Dishes → Washing Dishes \\

Other → Other & Chores → Chores & Clean → Other & Bed to Toilet → Bed to Toilet &  \\

& Dining Room Activity → Dining Room Activity & Wash Bathtub → Other & Laundry → Other & Computing → Computing \\

& Evening Meds → Other & Other → Other & Other → Other & Watching TV → Other \\

& Meditate → Meditate & & & Going Up → Other \\

& Other → Other & & & Going Down → Other \\

& & & & Eating → Eating \\

& & & & Cooking → Cooking \\

& & & & Other → Other \\
\bottomrule
\end{tabularx}
\end{table*}

\section{Training Settings}

We train the model for up to 30 epochs using the Adam optimizer and a ReduceLROnPlateau scheduler. Hyperparameters are selected via grid search over learning rates ([1e-2, 1e-3, 1e-4, 5e-5]) and weight decay values ([0, 1e-4, 1e-5]). For pretraining and finetuning, we also search over the same set of learning rates.  The optimal configuration was found to be a learning rate of 1e-4 and a weight decay of 0.

Both experimental settings use three-fold stratified cross-validation. In each fold, one subset is held out for testing, while the remaining two are split into training and validation sets. All virtual data is used during the pretraining phase without additional partitioning. We report the mean and standard deviation of accuracy, macro F1, and weighted F1 across the three folds, consistent with evaluation protocols in prior work.

\begin{figure*}
    \centering
    \begin{subfigure}[b]{0.35\textwidth}
        \centering
        \includegraphics[width=\linewidth]{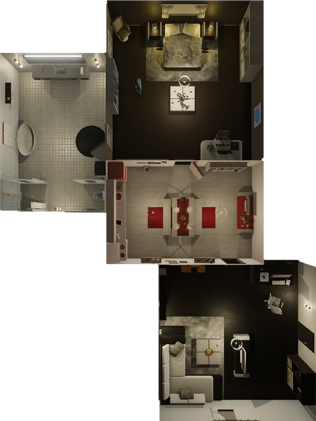}  
        \caption{VirtualHome layout}
        \label{fig:virtualhome}
    \end{subfigure}
    \hfill
    \begin{subfigure}[b]{0.45\textwidth}
        \centering
        \includegraphics[width=\linewidth]{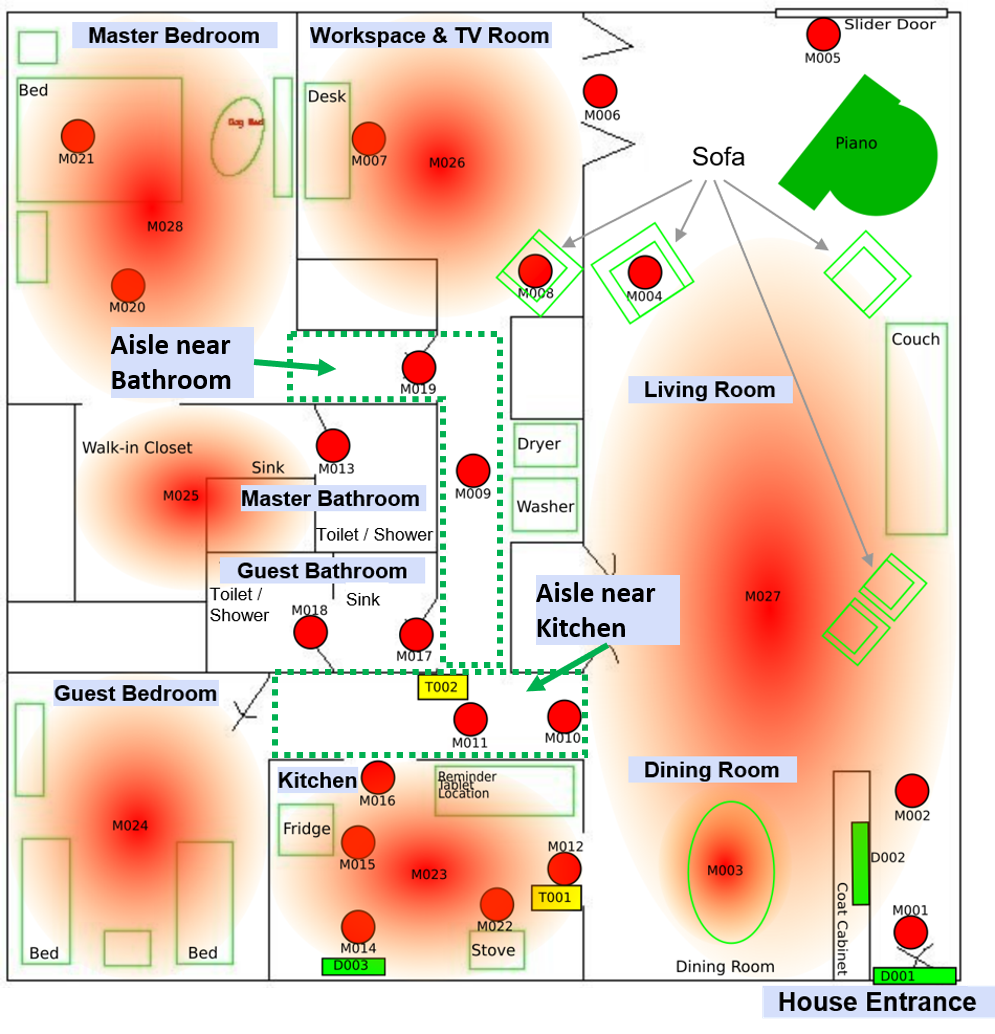}  
        \caption{Milan dataset layout}
        \label{fig:milan}
    \end{subfigure}
    \caption{Comparison between a VirtualHome environment and the Milan dataset. Virtual Home layout image from \cite{virtualhome2024} and Milan Layout taken from \cite{cook2012casas}}
    \label{fig:layout_comparison}
\end{figure*}

\section{Computing Infrastructure}
\label{computing infrastructure}
All experiments and simulations in this paper were conducted on a dedicated server equipped with an NVIDIA RTX A5000 GPU (24 GB VRAM), an Intel(R) Xeon(R) Gold 6342 CPU @ 2.80GHz, and 50 GB of system RAM. The server includes 48 physical cores and 96 logical processors across two CPU sockets, with Ubuntu 22.04.5 LTS (Jammy) as the operating system.
The software environment consists of Python 3.11.11, PyTorch 2.8.0.dev20250319+cu128, CUDA 12.8, and cuDNN 90800. Additional libraries include OpenCV, NumPy, and standard scientific computing packages.

\section{Discussion}
\label{discussion}
A key challenge in developing robust smart home-based HAR models is the lack of large-scale labeled datasets, further complicated by variations in home layouts, sensor setups, and individual behaviors. 
To address this, we introduce \textit{AgentSense}, a virtual data generation pipeline that uses LLMs to create diverse personas and routines, which are then executed in a simulated home instrumented with virtual sensors. 
This enables the generation of rich, diverse datasets that improve HAR model performance, particularly in settings where data is scarce. 
Models trained with our virtual data and minimal real data achieve performance on par with those trained on full datasets, demonstrating the potential of simulation-driven approaches to overcome data scarcity.

In the following section, we outline how insights gained from this work can inform future directions and detail the next steps in advancing our research.

\subsection{Bridging the Domain Gap Between Real and Virtual Homes}
\label{discussion-domainGap}
We generated virtual sensor datasets from 22 distinct environments within the VirtualHome simulator. While diverse, none of these virtual environments precisely replicate the home layouts found in the real-world HAR datasets used for evaluation. As illustrated in Figire \ref{fig:layout_comparison}, a comparison between a VirtualHome environment and the Milan dataset reveals noticeable differences—Milan, for instance, includes more rooms and unique spaces such as an aisle. These discrepancies in layout introduce a domain gap between the virtual and real data. Despite this, our results show that incorporating virtual data significantly improves HAR model performance, even in the presence of this domain mismatch.

One promising future direction is to leverage the control offered by simulation to replicate the specific layout of a target home when needed. This enables personalization of HAR models for specific homes at virtually no additional cost. By generating layout-matched virtual sensor data, the model can be better adapted to the unique movement patterns and spatial transitions of that home.

Consider, for example, a newly instrumented smart home with no prior sensor data. Traditionally, HAR model development in such settings requires a "cold phase" \cite{hiremath2022bootstrapping}—a period of passive observation during which user activities are manually annotated to create labeled training data. By extending \textit{AgentSense}, this phase could be bypassed entirely. A simple video walkthrough of the home can be used to reconstruct a 3D model of the environment using tools like Meshroom \cite{alicevision2021mesh}. This model can then be imported into the simulator to generate virtual sensor data that matches the actual home layout. 
With this virtual data, an initial HAR model can be trained and deployed from day one, enabling activity recognition without the need for manual labeling or real-world data collection -- thus supporting the transition to the maintenance procedure for HAR in homes \cite{hiremath2024maintenance}.

\subsection{Optimizing Sensor Placement Through Virtual Data Simulation for HAR}
\label{discussion-optimizeSensor}
We generated virtual sensor data using motion sensors placed according to predefined rules, as detailed in Section~\ref{sec: motion sensor}. However, the sensor placement can be fully customized within the virtual environment. This flexibility is particularly useful for identifying optimal sensor configurations when instrumenting a new home with ambient sensors.
Consider a scenario in which a new home is to be equipped with sensors to support recognition of a specific set of user-desired activities. Rather than deploying sensors directly and relying on trial-and-error in the real world, one can first simulate various sensor layouts in the virtual environment. By generating virtual sensor data under different configurations and evaluating the resulting HAR model performance, the optimal sensor layout can be identified—defined as the configuration that yields the highest activity recognition accuracy. This approach enables cost-effective, large-scale experimentation without the time and expense associated with real-world sensor deployment.

\subsection{Extending to Multi-modal Data Generation in Simulated Home Environments:}
As described in Section~\ref{sec: X-Virtualhome}, the VirtualHome simulator supports the generation of multiple camera-based modalities, including RGB images, depth maps, semantic segmentation, and pose data. Our extended version enhances this capability by adding support for multiple ambient sensor data streams that are time-synchronized with these visual modalities.

Among these, pose and video data are especially valuable for sensor-based HAR, as they allow the integration of existing Pose2IMU and Video2IMU methods~\cite{kwon2020imutube, xiao2021, Uhlenberg2022mesh, Xia2022, kwon2021complex, leng2023benefitgenerativefoundationmodels, Leng2024emotion, Hwang2024wheel} to generate synchronized virtual accelerometer and gyroscope signals. In future work, we plan to jointly simulate these sensor modalities, leveraging LLM-based activity generation to create richly annotated, multi-modal datasets. Such simulation of synchronized ambient and wearable sensor streams would enable more comprehensive activity analysis—combining the contextual awareness of ambient sensors with the fine-grained motion tracking offered by wearables, particularly for activities of daily living~\cite{arrotta2021marble}.

\subsection{Future Work}
\label{discussion-futureWork}
Several avenues for future work directly connect to the work we presented in this paper. 
In what follows, we outline some of these future avenues.

\subsubsection{Exploring Alternative Large Language Models for Virtual Data Generation:} 
In this work, we utilize GPT-4o-mini  or various language generation tasks, including persona creation, high-level routine synthesis, and decomposition into low-level action sequences. 
This model offers near-zero cost for generating the required sequences. 
It can be substituted with other open-source LLMs such as DeepSeek \cite{liu2024deepseek}, Claude \cite{anthropic2024claude35}, Gemini \cite{team2023gemini}, or Llama \cite{patil2024review, naveed2023comprehensive, touvron2023llama}.
In future iterations, we plan to investigate how different LLMs influence the generated action sequences and how these variations affect the resulting virtual data in simulation environments.

While our multi-stage prompting procedure and activity annotations rely on LLM-generated outputs, one could argue that these may not fully capture the nuances of real-world, situated home environments. 
In future work, we will include a formal evaluation of LLM outputs by comparing them against real resident inputs of annotation sourced via Amazon Mechanical Turk \cite{patel2022proactive}, as successfully demonstrated in prior studies.

\subsubsection{Exploring Additional Pretraining Methods with Virtual Data}
In this work, we utilized the state-of-the-art TDOST framework to assess the benefits of incorporating virtual data into the Human Activity Recognition (HAR) pipeline. 
Low-level action sequences generated by LLMs were fed into the VirtualHome simulator, and the resulting sensor event triggers were encoded using the TDOST representation. 
Evaluation was conducted by pre-training on virtual data, followed by fine-tuning with a combination of virtual and real data. 
This setup was benchmarked against the standard TDOST pipeline, which has demonstrated strong performance in generalizable, layout-agnostic HAR—where home layouts and activity patterns differ between source and target homes.
To examine whether virtual data enhances generalization, we directly compared TDOST alone with TDOST pre-trained on virtual data. 
The improved performance in the pre-trained variant highlights the utility of synthetic data for building robust HAR systems. 
In future work, we plan to expand this analysis by incorporating traditional baselines, for example, the use of CASAS features \cite{cook2012casas, alam2012review} and exploring other self-supervised pre-training strategies \cite{chen2024utilizing, chen2024enhancing, Haresamudram2022ssl, haresamudram2021contrastive, oord2018representation}.

\subsubsection{Enhancing Coverage and Diversity in HAR through LLM-Generated Activities:}
A notable advantage of incorporating LLMs in the simulation pipeline is their ability to generate broad spectrum of plausible human activities, many of which align with real-world behaviors yet remain underrepresented in existing HAR datasets. 
While such datasets are typically constrained by a fixed set of annotated labels, actual human routines are significantly more varied and context-dependent.
LLMs, trained on diverse and expansive textual corpora, can implicitly capture this behavioral richness and often produce activities that, though missing from benchmark datasets, mirror everyday human actions. 
In our generation process, we observed such instances, where the LLM surfaced realistic sequences absent from the labeled datasets -- highlighting limitations in traditional data collection pipelines, which are often bound by cost or activity taxonomies. 

Conversely, we also noted that certain activities commonly found in benchmark datasets, such as `Watch\_TV' or `Enter\_Home', were occasionally overlooked in the LLM-generated routines. 
This suggests areas where the open-ended simulation process could benefit from more targeted prompting 
In future work, we aim to explicitly include such activities to ensure alignment with established datasets and achieve fuller coverage of both common (occurring in benchmarked datasets) and underrepresented behaviors.

\end{document}